\pdfoutput=1
\documentclass[11pt]{article}

\usepackage[review]{EMNLP2023}

\usepackage{times}
\usepackage{latexsym}

\usepackage[T1]{fontenc}
\usepackage[utf8]{inputenc}

\usepackage{microtype}

\usepackage{inconsolata}
\usepackage{float}
\usepackage{booktabs}
\usepackage{graphicx}
\usepackage{amsmath}
\usepackage{amssymb}
\usepackage{enumitem}

\usepackage{microtype}

\usepackage{pifont}
\usepackage{adjustbox}
\usepackage{hyperref}
\usepackage{subcaption}
\usepackage{caption} 
\usepackage{fixltx2e}
\usepackage{multirow}
\usepackage{stfloats}

\title{SCO-VIST: Social Interaction Commonsense Knowledge-based\\ Visual Storytelling}

\author{Eileen Wang$^1$ 
        \quad Soyeon Caren Han$^{1,2}$\thanks{\ \ Corresponding author. caren.han@sydney.edu.au} 
        \quad Josiah Poon$^1$\\ 
    $^1$School of Computer Science, The University of Sydney \\
    $^2$School of Computing and Information Systems, The University of Melbourne\\ 
    \texttt{\{Eileen.Wang, Caren.Han, Josiah.Poon\}@sydney.edu.au} \\
  }

\begin{document}
\maketitle
\begin{abstract}
Visual storytelling aims to automatically generate a coherent story based on a given image sequence. Unlike tasks like image captioning, visual stories should contain factual descriptions, worldviews, and human social commonsense to put disjointed elements together to form a coherent and engaging human-writeable story. However, most models mainly focus on applying factual information and using taxonomic/lexical external knowledge when attempting to create stories. This paper introduces SCO-VIST, a framework representing the image sequence as a graph with objects and relations that includes human action motivation and its social interaction commonsense knowledge. SCO-VIST then takes this graph representing plot points and creates bridges between plot points with semantic and occurrence-based edge weights. This weighted story graph produces the storyline in a sequence of events using Floyd-Warshall's algorithm. Our proposed framework produces stories superior across multiple metrics in terms of visual grounding, coherence, diversity, and humanness, per both automatic and human evaluations. 
\end{abstract}

\section{Introduction}

Beyond interpreting the factual content of scenes with expressions, like image captioning, Visual Storytelling (VST) aims to conduct a human-like understanding of the idea of a sequence of images and generate more complicated visual scenarios with human-like textual expressions~\cite{huang2016visual}. In order to achieve this aim, the AI agent is required to model relationships between the images while remaining visually grounded, identify concepts that are implied (but not explicitly shown) in the images, as well as generate coherent, conversational language resembling how a human would tell a story in a social setting. 

Numerous past studies have employed encoder-decoder frameworks that first utilise a computer vision algorithm to extract image-specific features, which are then fed into a language generation model to decode the story \cite{DBLP:journals/corr/abs-1806-00738, Kim2018GLAC, jung2020hide, smilevski2018stories}. Although these methods can yield reasonable stories to some extent, they often lack common sense reasoning, thus producing stories that are "generic" sounding with limited vocabulary, and irrelevant to the images. To alleviate these issues, more recent approaches adopt content planning methods that try to explicitly predict textual concepts from the images via detecting objects in the image by using external knowledge data sources to identify implicitly related concepts \cite{chen2021commonsense, hsu2020knowledge, hsu2021plot, xu2021imagine}. Those external knowledge data sources mainly comprise taxonomic, lexical and physical relations, whereas human-like storytelling tends to use the social-aspect relations of everyday human experiences. Social-interaction relations comment on socially-triggered states and behaviours. It is crucial to gauge people's intentions and purpose and predict situationally-relevant human reactions and behaviours, which is directly aligned with the aim of human-like storytelling. 

This paper proposes a new social-interaction commonsense-enhanced VST framework, SCO-VIST, for producing human-like stories by interpreting socially-triggered situations and reactions. We introduce a three-stage commonsense enhanced framework that attempts to construct a reasonable plot of story events from the given image stream for story decoding. Stage 1 focuses on constructing a story graph representing causal and logical relationships between social interactions and events. Motivated by the idea that captions may already have embedded social commonsense within them, we first generate a caption for each image to literally capture the event depicted in the photo. Additionally, we further extract commonsense from external data related to social situations, interactions and behavioural responses (i.e. character's intentions, desires or needs). Each extracted caption and commonsense is thus considered a different event or plot point, and we connect the plot points (nodes) with causal ordering. In stage 2, we convert the story graph to be weighted by conducting a comprehensive analysis on different edge weight assignment methods based on semantic similarity between nodes and graph learning. Intuitively, this weighted story graph reflects the branching space of plausible event continuations where the edge weights indicate the likelihood of transition between connected plot points. Given the weighted story graph, the optimal storyline is the path of nodes that yields the largest sum of weights from the left to the right-most nodes in the graph. Therefore, Stage 3 negates the edge weights and employs Floyd-Warshall's shortest path search algorithm to extract the optimal sequence of story events which is later fed into a Transformer for story generation. The main contributions of this research are: 
% \begin{itemize}
%     \item We introduce a social-interaction commonsense enhanced VST framework that improves understanding of social situations and characters' feelings
%     \item We design a heterogeneous story graph and conduct a comprehensive analysis of the role of node and edge construction and learning over the visual storytelling dataset
%     \item We show that our model outperforms state-of-the-art when comparing automatic metrics, especially when analysing recently proposed metrics designed for VST
%     \item For robust evaluation, we also conduct human evaluation studies and demonstrate that our framework consistently and significantly outperforms several strong baselines
% \end{itemize}
\begin{itemize}
    \item We introduce a social-interaction commonsense enhanced VST framework that improves understanding of social situations and characters' feelings
    \item We design a heterogeneous story graph and conduct a comprehensive analysis of the role of node and edge construction and learning over the visual storytelling dataset
    \item We show that our model outperforms state-of-the-art when comparing automatic metrics, especially when analysing recently proposed metrics designed for VST 
    \item For robust evaluation, we also conduct human evaluation studies and demonstrate that our framework consistently and significantly outperforms several strong baselines.  
\end{itemize}

\section{Related Work}
Earliest works on VST consist of an encoder-decoder structure incorporated in an end-to-end model \cite{DBLP:journals/corr/abs-1806-00738, Kim2018GLAC, smilevski2018stories}. Recently, there has been increasing interest in reinforcement learning architectures which include a reward model to evaluate the generated stories \cite{hu2020makes, wang2018no}. However, the training process of such methods are inherently unstable. Other approaches first translate images to semantic scene graphs to capture image features and then employ Graph Convolutional Networks (GCN) to enrich regions and object representations \cite{han2020victr, hong2020diverse, wang2020storytelling}. Instead, we use literal text descriptions of images which can better explicitly represent the image contents. 

To promote more diverse stories, newer works have also used knowledge graphs to assist the storytelling process, allowing for richer stories capable of expressing imaginative concepts that are not explicitly shown in the image scene. Most of these methods involve querying ConceptNet \cite{speer2017conceptnet} with detected image objects or predicted key image concepts to find a set of related candidate concepts \cite{chen2021commonsense, xu2021imagine, yang2019knowledgeable}. While these methods show promising improvements in outputs, ConceptNet mainly comprises of taxonomic and physical relations, whereas our framework leverages commonsense that are more social-interaction focused and event-centred. Finally, most related to our work, recent studies try to form the story plot by first using external knowledge to connect concepts between images to reason about potential temporal relationships \cite{hsu2020knowledge, hsu2021plot, hsu2021stretch, wang2022me}. However, these methods often employ complex network architectures to iteratively predict subsequent events. We alleviate these complexities and present a simple yet effective approach for storyline construction.

\section{Method}
Figure \ref{fig: VIST_Method} depicts an overview of SCO-VIST's three stages. The following sections will describe each step in detail.

\subsection{Stage 1: Story Graph Construction} 
\textbf{Node Construction} The story graph contains 3 types of nodes: caption, commonsense and theme nodes. The caption nodes are obtained by using a pre-trained image captioning model to generate a textual description for each image in the photo sequence. That is, given the sequence of 5 images, captions $\{C_1, C_2,...C_5\}$ are generated where $C_i$ is the caption for the $i^{th}$ image. The intuition behind using captions is that literal descriptions of an image can provide more specific and accurate details about image contents compared to the raw visual features extracted from the image itself. Moreover, this step mimics how a human would tackle the VST task, as one would usually first consider what is visually represented in the image and its context before forming the premise of the story.

\begin{figure*}[t]
  \centering
  \includegraphics[width=1\linewidth]{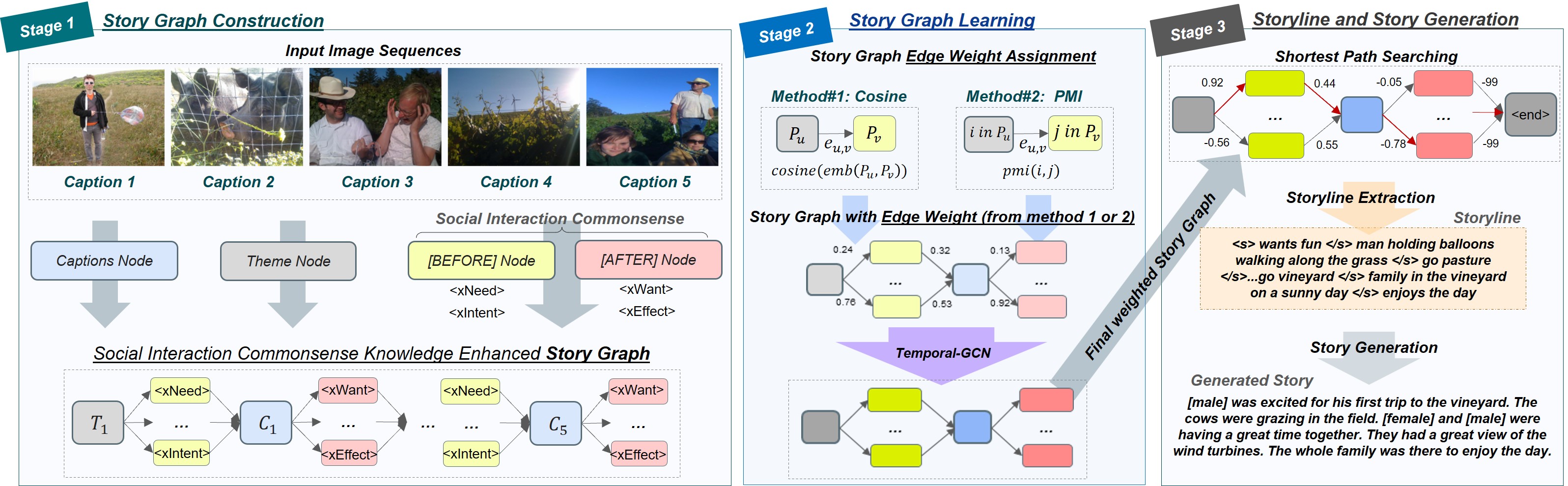}
  \caption{SCO-VIST's proposed framework. In Stage 1, the caption, theme and commonsense nodes are created and connected with causal ordering to form the story graph. In Stage 2, edge weights are assigned using cosine similarity or point mutual information and further refined through graph learning. Stage 3 takes the final story graph, negates the weights and constructs the storyline by finding the shortest path between the left and right-most node. The storyline is then fed to a Transformer for story generation. The corresponding detailed view of the final story graph for this example is depicted in Appendix \ref{Story Graph}.}
  \label{fig: VIST_Method}
\end{figure*}

%  \begin{figure*}[t]
%   \centering
%   \includegraphics[width=1\linewidth]{images/story_graph_farm.drawio.png}
%   \caption{The final directed story graph generated from Stage 2 with the additional dummy end node added in Stage 3. Grey and blue nodes are theme and caption nodes respectively. Yellow nodes are commonsense nodes from the \texttt{BEFORE} events group generated by the \texttt{xNeed} and \texttt{xIntent} relation while red nodes are the \texttt{AFTER} events commonsense nodes from the \texttt{xWant} and \texttt{xEffect} relation. Due to limited space, only the nodes corresponding to image 1, 2 and 5 are visualised and dotted lines are used to indicate nodes in the graph that are not displayed. The red highlighted arrows show the shortest path found by Floyd Warshall's algorithm where the caption nodes and commonsense nodes are taken in order to use as the storyline. For simplicity, edge weights are also not shown.}
%   \label{fig: storygraph}
% \end{figure*}

Next, we specifically focus on generating commonsense related to social interactions and dynamic aspects of everyday events. As such, Comet-ATOMIC2020 is utilised, a `\textit{neural knowledge model}' trained on the ATOMIC commonsense knowledge graph dataset \cite{hwang2021comet} which contains information on common human everyday experiences and mental states. Given a head/source phrase and relation (e.g. eat a cake \texttt{Intent}), Comet-ATOMIC2020 is capable of producing a tail phrase on-demand (e.g. celebrate birthday). Thus, out of the available 9 social interaction relations that Comet-ATOMIC2020 offers, we select 4 relations that primarily focus on causal and behavioural relationships: \texttt{xNeed}, \texttt{xIntent}, \texttt{xEffect} and \texttt{xWant}. More specifically, the \texttt{xNeed} relation indicates what event is needed to happen \textit{before} a following event occurs while the \texttt{xIntent} relation indicates a character's intention \textit{before} an action takes place. Conversely, \texttt{xEffect} are social actions that occur \textit{after} an event while \texttt{xWant} represents a character's postcondition desires \textit{after} an event. We append the 4 relation tokens to each caption phrase $C_{i}$ to provide as input for querying Comet-ATOMIC2020. Five commonsense inferences are generated per relation $r$, $\{ck_1^r, ck_2^r,...,ck_n^r\}$, resulting in 20 commonsense altogether for each caption.  The commonsense produced for each caption are then grouped into \texttt{BEFORE} and \texttt{AFTER} events. The \texttt{BEFORE} events category contains the knowledge extracted from the \texttt{xNeed} and \texttt{xIntent} relation while the \texttt{AFTER} events contains the \texttt{xEffect} and \texttt{xWant} commonsense. Finally, the theme nodes contain a sequence of concepts that represent the theme depicted in each image. We use Clarifai \footnote{\url{www.clarifai.com}}, a pretrained object and concept detector model capable of predicting 11,000 unique concepts. We extract a sequence of 20 concepts for each of the 5 images to create 5 theme nodes $\{T_{1}, T_{2}, T_{3}, T_{4}, T_{5}\}$. 
\\[7pt]
\noindent \textbf{Connecting Nodes} Let $CK_B=\{ck_{1}^r, ck_{2}^r,...,ck_{m}^r\}$ where $r\in\{\texttt{xNeed}, \texttt{xIntent}\}$ be the \texttt{BEFORE} commonsense inferences for caption $C_{i}$. Similarly, we denote $CK_A=\{ck_{1}^r, ck_{2}^r,...,ck_{m}^r\}$ where $r\in \{\texttt{xEffect}, \texttt{xWant}\}$ to be the \texttt{AFTER} commonsense inferences. To construct the story graph, we add directed edges between $T_{i}$ (the theme node for image $i$) and the commonsense nodes in $CK_B$. Each node in $CK_B$ is then connected to $C_i$ which is further connected to each node in $CK_A$. Finally, each node in $CK_A$ is connected with the theme nodes for the next image, $T_{i+1}$. Consequently, a directed acyclic graph $S_G$ representing the branching space of possible story events for each image stream is constructed as seen in Stage 1 of Figure \ref{fig: VIST_Method}. 

\subsection{Stage 2: Story Graph Learning} \label{weights} 
This stage conducts an analysis on the importance and role of each node in the story graph by converting $S_G$ into a weighted graph, $S_{G,weighted}$. Two main methods for edge weight assignment based on semantic similarity is experimented with and weights are further refined with graph learning. 
\\[7pt]
\noindent \textbf{Cosine Similarity} Firstly, we use the cosine similarity between plot points as an indicator of their level of association. Given connecting nodes $u$ and $v$ which contain words or a phrase denoted by $P_u$ and $P_v$ respectively, we convert $P_u$ and $P_v$ to a sentence embedding using a pretrained transformer model. The cosine similarity score between the two embeddings at node $u$ and $v$ is then simply assigned to their connecting edge $e_{u,v}$. 
\\[7pt]
\noindent \textbf{Pointwise Mutual Information (PMI)} The second method computes the PMI between each pair of words in $P_u$ and $P_v$ where a high PMI implies high semantic correlation between words. Formally, the PMI between word $i$ in $P_u$ and word $j$ in $P_v$ is:
\begin{equation}
    \text{PMI}(i,j) = \text{log}\frac{p(i,j)}{p(i)p(j)}
\end{equation}

\noindent Here, $p(i,j) = \frac{\text{\#}S(i,j)}{\text{\#}S}$, $p(i) = \frac{\text{\#}S(i)}{\text{\#}S}$ and $p(j) = \frac{\text{\#}S(j)}{\text{\#}S}$ where $\text{\#}S(i)$ is the number of sentences in the corpus that contain word $i$, $\text{\#}S(i,j)$ is the number of sentences that contain both words and $\text{\#}S$ is the total number of sentences in the corpus. Finally, a normalized version of the PMI score is calculated: 
\begin{equation}
    \text{NPMI} = \frac{\text{PMI}}{-\text{log}(p(i,j))}
\end{equation}
The final weight assigned to $e_{u,v}$ is the maximum NPMI score out of all scores calculated from the possible word pair combinations. 
\\[7pt]
\noindent \textbf{Graph Learning} We further refine the cosine or PMI-weighted story graph through graph learning. Specifically, the weighted graph is fed into a Temporal Graph Neural Network (TGCN). Such networks combine the advantages of GCNs and Recurrent Neural Networks to learn the graph's complex topological structure as well as its temporal changes. We use an implementation of the Gated Graph Convolution Long Short Term Memory Layer \cite{taheri2019predictive} which encodes the graph and yields embeddings for each node. We then extract the 5 embeddings from the caption nodes and feed them through the BART Transformer \cite{lewis2020bart} to decode the story. The TGCN and Transformer are trained end-to-end to minimise the cross-entropy loss:

\begin{equation}
\label{ce-loss}
    L(\theta) = -\sum_{t=1}^{T} \text{log}(p_{\theta}(y_{t}^{*}|y_{1}^{*},...,y_{t-1}^{*}))
\end{equation}
where $\theta$ is the parameters of the model, $y^{*}$ is the ground-truth story and $y_{t}^{*}$ denotes the $t$-th word in $y^{*}$. Finally, we extract the learnt node embeddings and compute the cosine similarity between the embeddings of each pair of connected nodes to obtain the edge weight in between.

\subsection{Stage 3: Storyline and Story Generation} 
\textbf{Storyline Extraction} Given $S_{G,weighted}$, we consider the optimal storyline as the path from the left-most node to the right-most node that produces the highest sum of weights. To find this path, we negate each weight in $S_{G,weighted}$ and add a dummy end node $D_E$ which is connected with the right-most nodes in $S_{G,weighted}$ with an edge weight of -99. An example of the final graph is depicted in Appendix \ref{Story Graph}. Floyd–Warshall's algorithm \cite{floyd1962algorithm} is then adopted to find the shortest path starting from $T_1$ to $D_E$ to produce the storyline containing a sequence of events $e_1,...,e_L$ taking only the caption and commonsense nodes.
\\[7pt]
\noindent \textbf{Story Generation} The last stage consists of decoding the story. We separate each event $e_{i}$ using a separator token $\texttt{</s>}$. The events are then fed through BART for story generation which we train with the cross-entropy loss from Equation \ref{ce-loss}. 

\section{Evaluation Setup\footnote{Implementation details can be found in Appendix \ref{implementation_details}}}
\textbf{Data} The Visual Storytelling Dataset (VIST) \cite{huang2016visual} consists of 210,819 unique images obtained from Flickr albums. The dataset is split into training/validation/testing with 8,031/998/1,011 albums where each album contains a set of similar image sequences with each sequence made up of 5 photos. Each album also has 5 human written stories where each story is usually comprised of one sentence per image. The unique number of stories in the training, validation and testing set is 40,155, 4,990 and 5,055 respectively.

\subsection{Baseline Models}
% We compare our model with 6 state-of-the-art baselines: 1) \textbf{AREL} \cite{wang2018no}, 2) \textbf{GLACNet} \cite{Kim2018GLAC}, 3) \textbf{KG-Story} \cite{hsu2020knowledge}, 4) \textbf{ReCo-RL} \cite{hu2020makes}, 5) \textbf{PR-VIST} \cite{hsu2021plot}, and 6) \textbf{TAPM} \cite{yu2021transitional}. Details of each model is outlined in Appendix \ref{baseline models}.
We compare ours with 6 state-of-the-art baselines. 
\begin{enumerate}
    \addtolength\itemsep{-2.5mm}
    \item \textbf{AREL} \cite{wang2018no} adopts an inverse reinforcement learning (RL) approach trained in an adversarial manner with a CNN-based reward model.
    \item \textbf{GLACNet} \cite{Kim2018GLAC} is an end-to-end model that combines both local and global attention mechanisms on the image features.
    \item \textbf{KG-Story} \citep{hsu2020knowledge} attempts to enrich stories by leveraging external knowledge bases like Visual Genome \cite{krishna2017visual} and OpenIE \citep{pal2016demonyms}. For story generation, a Transformer model is used.
    \item \textbf{ReCo-RL} \cite{hu2020makes} proposes another RL method with composite rewards designed to target the \textit{relevance}, \textit{coherence} and \textit{expressiveness} criteria of VST. 
    \item \textbf{PR-VIST} \cite{hsu2021plot} is a newer model where similar to ours, attempts to link nouns together with verb relations extracted from Visual Genome and VIST to form a story graph. The optimal storyline is then extracted using UHop \cite{chen2019uhop}.
    \item \textbf{TAPM} \cite{yu2021transitional} introduces an auxiliary training task to harmonise the language generator and visual encoder before optimising the target objective. The task proposes to minimise the `sequential coherence loss' which aims to enforce text representations to predict surrounding visual representations within a closed neighbourhood. 
\end{enumerate}

\subsection{Ablation Study Models}\label{ablation_section}
We also conduct ablation studies to compare different variants of our proposed model: 
1) \textbf{SRL-caption}: A story graph is not created and the 5 image captions are used as the storyline, 
2) \textbf{SRL-pmi/cosine}: The storyline is extracted from the story graph using weights obtained from the cosine similarity or PMI approach.
3) \textbf{TGCN/TGCN-SRL}: TGCN-cosine/pmi is an end-to-end model where the story graph is fed to the TGCN and node embeddings are then inputted into BART for story decoding. The story graph input uses weights obtained from either the cosine or PMI approach. TGCN-SRL-cosine/pmi further uses the trained TGCN to extract the node embeddings and their similarities are then used to refine the story graph weights for storyline and story generation.
% \raggedbottom
% \begin{itemize}[leftmargin=0.5cm]
%     \addtolength\itemsep{-2.2mm}
%     \item \textbf{SRL-caption}: A story graph is not created and the 5 image captions are used as the storyline. 
%     \item \textbf{SRL-pmi/cosine}: The storyline is extracted from the story graph using weights obtained from the cosine similarity or PMI approach.
%     \item \textbf{TGCN/TGCN-SRL}: TGCN-cosine/pmi is an end-to-end model where the story graph is fed to the TGCN and node embeddings are then inputted into BART for story decoding. The story graph input uses weights obtained from either the cosine or PMI approach. TGCN-SRL-cosine/pmi further uses the trained TGCN to extract the node embeddings and their similarities are then used to refine the story graph weights for storyline and story generation.
% \end{itemize}

\subsection{Automatic Metrics}
Numerous past literature have shown that traditional automatic metrics like BLEU correlate poorly with human judgement and are unreliable for evaluating VST \cite{wang2018no, hsu2019visual}. These metrics mainly focus on comparing $n$-gram similarity between hypothesis and references, thus are insufficient for evaluating open-ended text generation tasks like storytelling, where there are multiple plausible outputs for the same input which are not fully reflected in the references. Therefore, we focus on metrics specifically designed for \textit{`open ended text generation}' which consider the plausibility of diverse outputs. The first is RoViST \cite{wang-etal-2022-rovist}, an unreferenced metric set for VST consisting of three scores that target three criteria: visual grounding (RoViST-VG), coherence (RoViST-C) and no redundant repetition of concepts/words (RoViST-NR). An overall single score (RoViST) can be calculated by averaging RoViST-VG, C and NR. In addition to RoViST, we consider other learnt \textit{`unreferenced'} metrics such as Perplexity and the storytelling metric, UNION \cite{guan2020union} which assigns a score based on important story criteria like coherence, no conflicting logic and non-repeating plots. Finally, for completeness and maintaining consistency with other works, we further compute reference-based metrics including the classic ROUGE-L \cite{lin2004rouge}, METEOR \cite{banerjee2005meteor}, CIDEr \cite{vedantam2015cider} and SPICE \cite{anderson2016spice}. For analysing semantic similarity, the BERT-based metric BLEURT \cite{sellam2020bleurt} is further adopted as well as the embedding-based metric, MoverScore \cite{zhao2019moverscore}.  %\footnote{\url{https://github.com/google-research/bleurt}}

%  As we find that RoViST-NR tends to award higher scores for extremely short sentences which have lower chance of repeating words, we slightly modify this score to penalize short sentences using a similar approach to BLEU's brevity penalty. Specifically, we multiply the original RoViST-NR score by $e^{1-\frac{r}{c}}$ where $c$ is the candidate story's length and $r$ is the average reference story length in the corpus if $c \leq r$. 

\begin{table*}[t!]
  \centering
  
  \resizebox{1\linewidth}{!}{
  \begin{tabular}{@{}l||ccc|c|ccccc|c|cc@{}}
    \toprule
     \textbf{Model} & \textbf{RoViST-VG} & \textbf{RoViST-C} & \textbf{RoViST-NR}  & \textbf{RoViST (R)}  & \textbf{SPICE (S)} & \textbf{BLEURT (B)} & \textbf{MoverScore (M)} & \textbf{UNION (U)} & \textbf{Perplexity} & \textbf{R+S+B+M+U} & \textbf{Story Len.} \\ \midrule
    AREL (\cite{wang2018no}) & 66.2 & 57.1 & 83.4 & 68.9 & 9.0 & 32.6 & 55.1 & 17.1 & 15.3 & 182.7 & 44.8 \\
    GLACNet \cite{Kim2018GLAC} & 61.6 & 68.6 & 95.1  & 75.1 & 7.0 & 33.5 & 54.9 & 75.9 & 24.6  & 246.3 & 35.2 \\
    KG-Story \cite{hsu2020knowledge} & 58.7 & 65.1 & \textbf{99.9} &  74.6 & 7.2 & 32.3 & 54.9 & 65.8 & 46.1 & 234.8 & 32.3\\
    ReCo-RL \cite{hu2020makes} & 67.8 & 57.3 & 91.9  & 72.3 & 11.2 & 31.9 & 55.4 & 23.8 & 28.3 & 194.6 & 49.3  \\ 
    PR-VIST \cite{hsu2021plot} & 70.0 & 60.4 & \underline{96.1} & 75.5 & 9.6 & 31.0 & 54.7 & 30.3 & 42.3 & 201.1 & \underline{52.2} \\
    TAPM \cite{yu2021transitional} & \underline{70.3} & 67.0 & 90.5 & 75.9 & 9.9 & 33.4 & 55.6 & 56.0 & 18.3 & 230.8 & 51.2 \\
    \hline
    SRL-caption & 65.2 & 73.9  & 91.4  & 76.8 & 6.1 & 31.7 & 53.3 & 76.5 & 16.0 & 244.5 & 49.7 \\
    SRL-cosine & 69.6 & 72.1  & 91.9 & \underline{77.9} & 11.2 & 34.6  & \textbf{56.0} & 78.8 & 15.1 & 258.4 & 48.0\\
    SRL-pmi & \textbf{70.4} & 72.8  & 91.6  & \textbf{78.3} & \textbf{11.5} & 34.7 & \textbf{56.0} & 75.9 & \underline{14.7} & 256.3 & 51.2\\
    TGCN-SRL-cosine & \underline{70.3} & 72.3 & 90.5 & 77.7  & 10.9 & \textbf{34.9} & \textbf{56.0} & 84.0 & 14.9 & \textbf{263.4} & \textbf{52.3} \\
    TGCN-SRL-pmi & 69.0 & 71.9  & 91.6 & 77.5 & \underline{11.2} & \underline{34.7} & \textbf{56.0} & 80.6 & \textbf{13.6} &  259.9 & 51.5  \\
    TGCN-cosine & 65.7 & \underline{75.5}  & 91.8 & 77.6 & 9.2 & 33.9 & 55.6 & \underline{84.3} & 16.5 & \underline{260.6} & 39.1\\
    TGCN-pmi & 65.7 & \textbf{75.9}  & 91.3 & 77.6 & 9.4 & 33.8 & \underline{55.7} & \textbf{87.0} & 15.5 & \textbf{263.4} & 40.5 \\
    \bottomrule
  \end{tabular}}
  \caption{Automatic metrics and average story length (Story Len.) for the 6 baselines vs. our 7 model variants.}\label{automatic_metrics}
\end{table*}

\subsection{Human Evaluation} 
We finally conduct human evaluation studies and create 3 surveys where each survey conducts a pairwise comparison between our model and a baseline. In the survey, participants are given 100 randomly selected unique photo sequences from the test data (same sequences are used for each survey) and the corresponding generated story from our model and the baseline. They are then asked to choose which of the two stories are better based on 3 criteria: 1) \textbf{Visual Grounding}: the generated story must relate to concepts depicted in the image sequence, 2) \textbf{Coherence}: story sentences need to flow while remaining logical and topically consistent, and 3) \textbf{Non-Redundancy}: sentences are diverse and there are no unnatural-sounding repetition of words/phrases in the story. A final question also asks the annotator to choose which story is better out of the two based on their opinion. 15 respondents (5 per survey) were recruited where each participant answered 400 questions, resulting in 6000 instances collected in total. 

% \begin{enumerate}
%     \addtolength\itemsep{-2.2mm}
%     \item \textbf{Visual Grounding}: the generated story must relate to concepts depicted in the image sequence.
%     \item \textbf{Coherence}: story sentences need to flow while remaining logical and topically consistent.
%     \item \textbf{Non-Redundancy}: sentences are diverse and there are no unnatural-sounding repetition of words/phrases in the story.
% \end{enumerate}

\section{Results}

\subsection{Overall Performance} 
Table \ref{automatic_metrics} summarises several metrics for the 6 baselines and for the 7 different variations of SCO-VIST. After filtering out broken images in the test set and missing stories from the baseline models, a sample of 890 albums was used to calculate these metrics. Considering our best model based on the visual storytelling metric RoViST (SRL-pmi), RoViST-VG performs on par with the more recent baselines and significantly outperforms in RoViST-C when considering all our 7 model variants. RoViST-NR however underperforms, but we strongly emphasize that this is most likely attributed to the short story lengths which have a lower chance of repeating words as can be seen by KG-Story which has a repetition score of 99.9 but average story length of only 32. 

Furthermore, studies in \citet{wang-etal-2022-rovist} emphasized that humans considered coherence to play the most significant role when judging a story, followed by visual grounding and non-redundancy. Nevertheless, our models still achieve noticeably better performance than the baselines when comparing the overall RoViST with SRL-pmi considered as the best model as it achieved a good balance of high scores across RoViST-VG, C and NR.

\raggedbottom
% Furthermore, we note that studies in \cite{wang-etal-2022-rovist} emphasized that humans considered coherence to play the most significant role when judging a story, followed by visual grounding and non-redundancy. Therefore, while PR-VIST outperforms our models in RoViST-NR, it has much lower coherence which significantly undermines their output's quality.

Although classic automatic metrics are known to correlate poorly with human judgement for VST, it is still noteworthy to analyse them in conjunction with RoViST. Hence, ROUGE-L, METEOR and CIDEr are shown in Table \ref{classic automatic} of where we observe that SRL-pmi resulted in lower scores. This could be due to our model using knowledge from COMET-Atomic2020 to enrich lexical diversity which results in lower performance in $n$-gram matching between the generated and reference stories. However, SRL-pmi still outperforms the baselines when comparing less classic metrics like SPICE which focuses on semantic propositional content, BLEURT which is based on semantic meaning and slightly on MoverScore which compares distances of word embeddings between reference and hypothesis stories. The unreferenced metrics for evaluating open-ended text generation, Perplexity and UNION also show significant improvements. Most noticeably, UNION which scores based on coherence, conflicting logic and chaotic scenes is able to reach an upper bound score of 87.0 with TGCN-pmi.

Finally, to gain a better overview of the overall performance, we sum RoViST, SPICE, BLEURT, MoverScore, and UNION and present the scores in the R+S+B+M+U column. When comparing the sum, the best performing models were the TGCN methods with TGCN-SRL-cosine and TGCN-pmi producing the highest scores.

\begin{table}[t]
  \centering
  \resizebox{0.8\linewidth}{!}{
  \begin{tabular}{@{}l||ccc}
    \toprule
     \textbf{Model} & \textbf{ROUGE-L} & \textbf{METEOR} & \textbf{CIDEr}  \\ \midrule
    AREL & \textbf{29.9} & 35.2 & 9.1 \\
    GLACNet & 27.2 & 33.5 & 4.4 \\
    KG-Story & 25.2 & 31.5 & 3.8  \\
    ReCo-RL & 29.3 & \textbf{35.9} & \textbf{11.9} \\
    PR-VIST & 26.1 & 31.4 & 7.6 \\  
    TAPM & 21.7 & 27.0 & 4.5 \\ \hline
    SRL-pmi & 22.1 & 27.5 & 5.9  \\
    \bottomrule
  \end{tabular}}
  \caption{Classic $n$-gram metrics for our top model, SRL-pmi vs. the 6 baselines.} \label{classic automatic} 
\end{table}

%When also comparing RoViST + SPICE + BLEURT in Table \ref{automatic_metrics}, SRL-pmi achieves a 6.6\% relative improvement over the best baseline (PR-VIST).

\subsection{Ablation Study} 
To analyse the effect of the storyline extraction stage and different edge weight assignment methods, an ablation study was conducted to compare the 7 different variations described in Section \ref{ablation_section}. We first compare just using the 5 captions (SRL-caption) as the storyline versus extracting the storyline from the commonsense story graph (SRL-cosine/pmi, TGCN-SRL-cosine/pmi). Surprisingly, competitive RoViST-C and NR scores was achieved from SRL-caption but underperforms substantially in the VG criteria. Additionally, SPICE, BLEURT, MoverScore, UNION and Perplexity were considerably worse. This implies that captions alone have sufficient commonsense embedded in them and can be useful features for generating plausible stories. However, the VG aspect can be further enhanced by exploiting extra social commonsense from external data.

Moreover, the TGCN-cosine/pmi approach consisting of the end-to-end model with a TGCN combined with the Transformer evidently produces lower RoViST-VG and NR compared to the SRL methods. SPICE, BLEURT, MoverScore and Perplexity scores were also mostly less optimal. This suggests that feeding the node embeddings into the Transformer for story decoding is not as good as extracting the storyline and explicitly using the words as input which can provide more fine-grained details about the image contents for generating richer stories. However, TGCN-cosine/pmi noticeably yielded the best RoViST-C scores out of the 7 methods ($>75$). This could be attributed to the shorter outputs as it is often easier to stay coherent with shorter generic sentences. 

Finally, it is interesting to note that higher UNION scores were obtained for all TGCN methods when compared to not using the TGCN. It is hypothesised  that incorporating learnt temporal information in the node embeddings implicitly via TGCN training perhaps resulted in more logical stories, thus improving the UNION score. 

% When comparing different edge weight assignment methods, all methods performed similarly with SRL-pmi slightly outperforming the others. This indicates that for optimal performance, it is not necessary to use the TGCN for weight refining.

\subsection{Visualising Diversity} 
 \begin{figure}[H]
  \centering
  \includegraphics[width=1\linewidth]{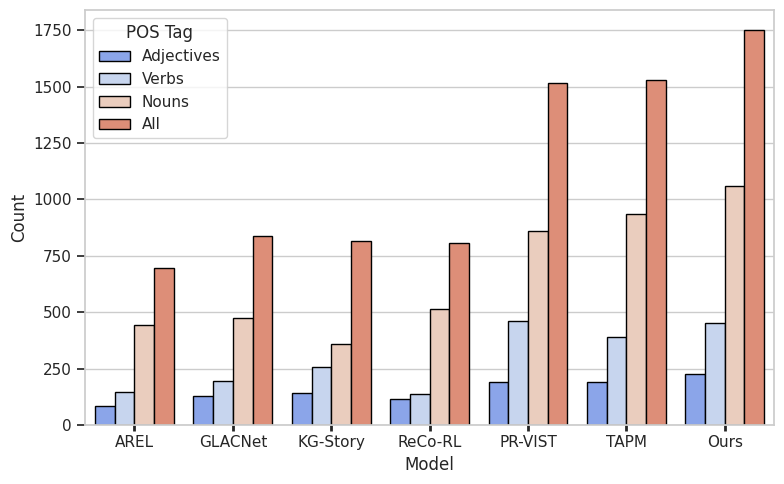}
  \caption{Count of unique unigrams for different part-of-speech (POS) tags for our proposed SRL-pmi vs. the 6 state-of-arts baselines.}
  \label{fig: bar chart}
\end{figure}

We visualise the number of distinct unigrams, nouns, verbs and adjectives outputted by SRL-pmi versus the 6 baselines. Figure \ref{fig: bar chart} illustrates that our model can produce significantly more unigrams overall especially when comparing nouns, suggesting that leveraging social interaction commonsense and the captions can generate richer and diverse sentences with more novel expressions. 

\subsection{Qualitative Analysis} 

\begin{figure}[H]
  \centering
  \includegraphics[width=1\linewidth]{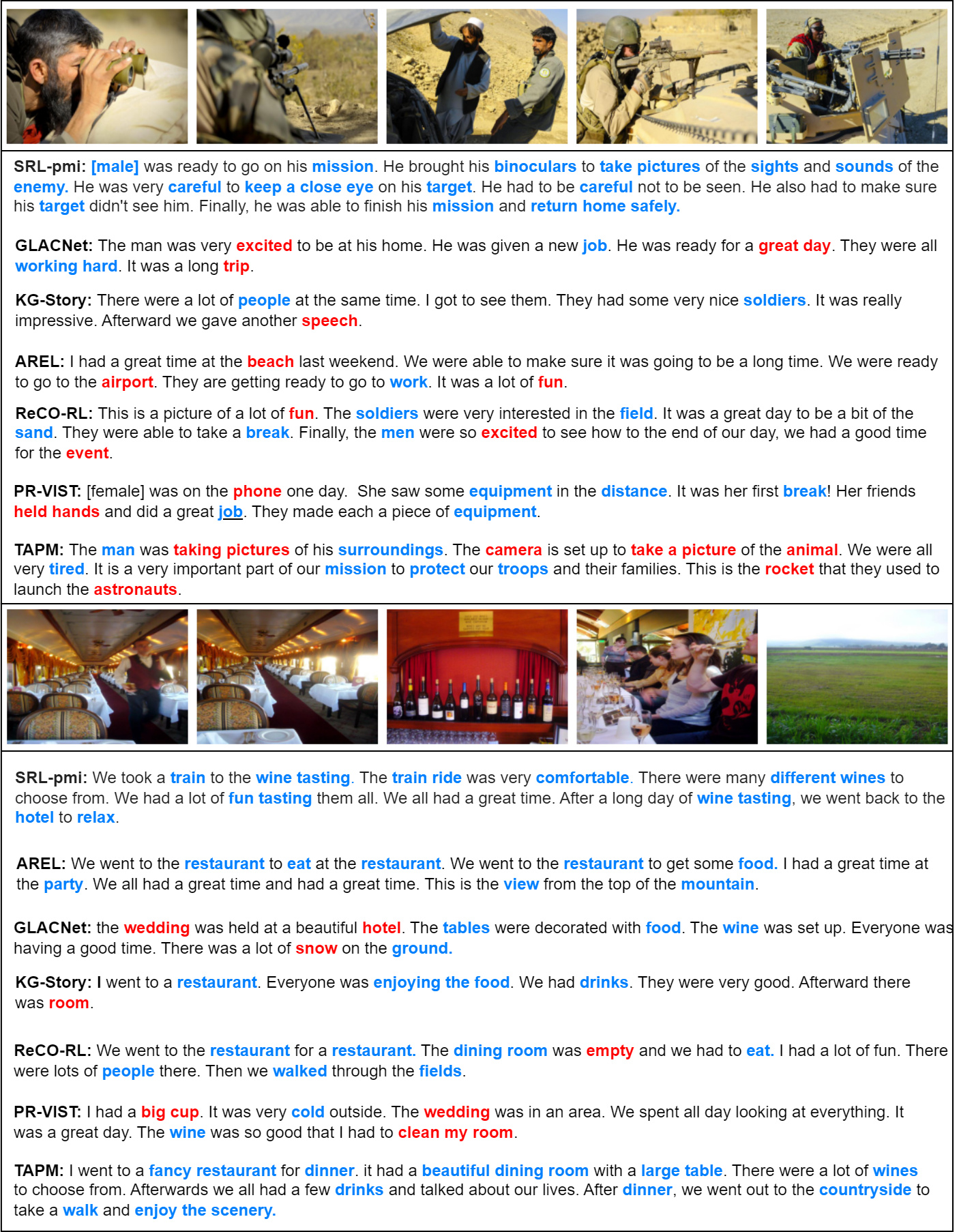}
  \caption{Generated stories for our SRL-pmi model versus the 6 baselines models. Blue/red words represent concepts relevant/irrelevant to the image sequence.}
  \label{fig: qual_analysis}
\end{figure}

To evaluate our model qualitatively, we show examples of generated stories from SRL-pmi versus the 5 baselines. Figure \ref{fig: qual_analysis} illustrates that our model generates stories that are clearly more visually grounded. For instance, ReCo-RL in the first example mentions several irrelevant phrases like `\textit{lot of fun}' while KG-Story incorrectly mentions `\textit{gave another speech}' in the last sentence. On contrary, our model's stories are more detailed and less generic such as the phrase, `\textit{ready to go on his mission}' and `\textit{sights and sounds of the enemy}', thus highlighting the effectiveness of using captions and social commonsense to capture events depicted and implied by the images. By not solely relying on visual features and using literal descriptions and commonsense to construct storylines as input, our stories are also consequently more coherent and natural-sounding. Taking the last sentence from AREL in the second story as an example, `\textit{This is the view from the top of the mountain}' sounds abrupt and is unrelated to the previous generated sentences. Conversely, our story is capable of capturing the changes between images while maintaining a strong focus on the topic of `\textit{wine tasting}'.
\raggedbottom

\begin{table*}[t]
    \renewcommand{\arraystretch}{1}
  \centering
  \resizebox{0.9\linewidth}{!}{
  \begin{tabular}{@{}l||cccc|cccc|cccc}
    \toprule
     & \multicolumn{4}{|c|}{\textbf{All Stories}} & \multicolumn{4}{|c|}{\textbf{Event-based}} & \multicolumn{4}{|c}{\textbf{Object-based}}  \\\hline
     \textbf{Criteria} & \textbf{Ours} & \textbf{AREL} & \textbf{Tie} & \textbf{Agree} & \textbf{Ours} & \textbf{AREL} & \textbf{Tie}  & \textbf{Agree} & \textbf{Ours}  & \textbf{AREL} & \textbf{Tie} & \textbf{Agree} \\ \midrule
    Visual Grounding & 88.0\% & 6.6\% & 5.4\% & 0.64 & 86.5\% & 8.5\% &  5\% & 0.60 & 96.9\% & 2.5\% & 0.6\% & 0.71 \\
    Coherence & 90.0\% & 4.8\% & 5.2\%  & 0.70 & 88.2\% & 5.3\% & 6.5\% & 0.66 & 93.8\% & 3.7\% & 2.5\% & 0.82 \\
    Non-Redundancy & 83.6\% & 3.0\% & 13.4\%  &  0.56 & 82.4\% & 3.2\% & 14.4\% & 0.54 &  86.3\% & 2.4\% & 11.3\% & 0.60 \\
    Overall & 93.4\% & 4.4\% & 2.2\% & 0.78 & 91.8\% & 5.3\% & 2.9\%  & 0.74 & 96.9\% & 2.5\% & 0.6\% & 0.88 \\ \hline
    \textbf{Criteria} & \textbf{Ours} & \textbf{ReCo-RL} & \textbf{Tie} & \textbf{Agree} &  \textbf{Ours} & \textbf{ReCo-RL} & \textbf{Tie}  & \textbf{Agree} &  \textbf{Ours}  & \textbf{ReCo-RL} & \textbf{Tie} & \textbf{Agree}   \\ \midrule
    Visual Grounding & 82.2\%  & 10.0\% & 7.8\% & 0.49 &  82.3\% & 10.6\%  & 7.1\% & 0.49 & 81.9\% & 8.7\% & 9.4\% & 0.48\\
    Coherence & 93.4\% & 4.2\% & 2.4\% & 0.78 & 94.7\% & 3.8\%  & 1.5\%  & 0.81 & 90.6\% & 5\% & 4.4\%  & 0.70 \\
    Non-Redundancy & 71.6\% & 11.0\% & 17.4\% & 0.30 & 72.3\% & 12.4\% & 15.3\% & 0.31 & 70.0\% & 8.1\% & 21.9\% & 0.28 \\
    Overall & 92.2\%  & 5.0\% & 2.8\% & 0.75 & 93.8\% & 4.1\% & 2.1\% & 0.79 & 88.8\% & 6.8\% & 4.4\%  & 0.66\\ \hline
    \textbf{Criteria} & \textbf{Ours} & \textbf{PR-VIST} & \textbf{Tie} & \textbf{Agree} &  \textbf{Ours} & \textbf{PR-VIST} & \textbf{Tie}  & \textbf{Agree} &  \textbf{Ours}  & \textbf{PR-VIST} & \textbf{Tie} & \textbf{Agree}   \\ \midrule
    Visual Grounding & 78.8 \%  & 17.2 \% & 4.0 \% & 0.52 & 79.1\% & 16.2\%  & 4.7\% & 0.52 & 78.1\% & 19.4\% & 2.5\% & 0.52 \\
    Coherence  & 77.8\% & 18.4\% & 3.8 \% & 0.44 & 79.4\% & 17.9\%  & 2.7\% & 0.47 & 74.4\% & 19.3\% & 6.3\% & 0.35 \\
    Non-Redundancy & 63.0\% & 24.6\% & 12.4\% & 0.28 & 64.4\% & 22.4\% & 13.2\% & 0.29 & 60.0\% & 29.4\% & 10.6\% & 0.23\\
    Overall  & 78.0\%  & 16.4\% & 5.6\% & 0.46 & 78.5\% & 16.2\% & 5.3\% & 0.48  & 76.9\% & 16.8\% & 6.3\%  & 0.43 \\
    \bottomrule
  \end{tabular}}
  \caption{Pairwise comparison between SRL-pmi with AREL, ReCo-RL and PR-VIST across the visual grounding, coherence, and non-redundancy criteria for all stories (500 instances) and when separated into event-based (340 instances) and object-based (160 instances) story categories. The `Agree' column shows the Fleiss' Kappa results. } \label{human_eval} 
\end{table*}

\subsection{Human Evaluation: Pairwise Comparison}
Table \ref{human_eval} reports the results of the pairwise comparison between SRL-pmi with AREL, ReCo-RL and PR-VIST. The last column (`Agree') represents results from the Fleiss' kappa test used to assess inter-rater consistency \cite{fleiss1971measuring}. Agreement scores in the range [0.21, 0.40], [0.41, 0.60] and [0.61, 0.80] means fair, moderate and strong agreement between multiple annotators respectively. 

When analysing all stories (`All Stories' sub-table), our generated stories evidently outperform the baselines by a large margin. All percentages in the first column are over 63\%, indicating that the majority of annotators selected our story to be better across all criteria. Moreover, when comparing the `Overall' criteria which asked evaluators to choose the better story, over 78\% of the responses reported our stories to be better with the Fleiss' kappa test result showing a moderate to strong level of agreement between annotators. We believe the higher votes for the visual grounding criteria for our model is due to our method incorporating relevant social-interaction commonsense. Additionally, our constructed storyline is able to reflect the causal events implied by the image stream, resulting in improved story coherence and less repetition. 

\subsection{Human Evaluation: Story Categories} \label{Human Evaluation: Story Categories}
\begin{figure}[t]
  \centering
  \includegraphics[width=0.95\linewidth]{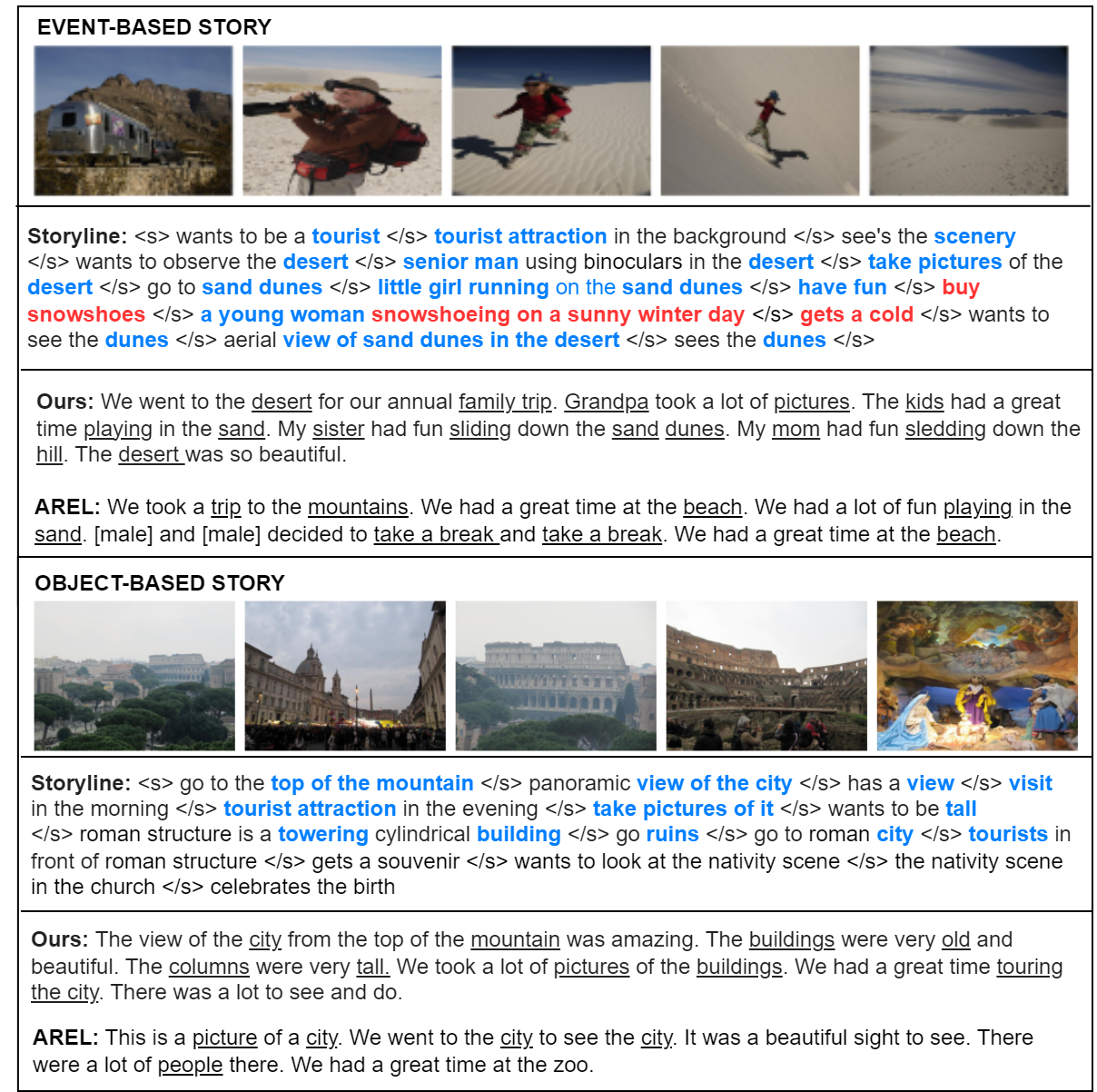}
  \caption{AREL vs. SRL-pmi for an event-based story (above) and object-based story (below). Blue words indicate concepts implicitly or explicitly used in the generated story while red represents irrelevant concepts. Underlined words in the story represent concepts relevant to the image stream.}
  \label{fig: human_eval_qual}
\end{figure}

We analyse the human evaluation results by categorising the stories into `event-based' and `object-based'. Event-based stories refer to image streams that focus on people performing actions and there is a clear transition of events between images. Object-based consists of images that mostly picture landscapes and objects. Such instances have no clear event in the image, and thus require more imagination when creating the story. An example of an event-based story is the top sequence in Figure \ref{fig: human_eval_qual} where we can clearly see a man taking a photo and a girl running and sliding across the sand. Conversely, the second example is object-based as a majority of the images depict scenery and buildings. It is harder to generate a story from this input as the first 4 images are extremely similar while the last image is totally different. 

Observing the last two sub-tables of Table \ref{human_eval}, the first baseline AREL shows lower percentages and ties for object versus event-based stories. As AREL purely relies on generating stories from the visual features, it fails to create coherent output particularly when consecutive images are similar. We qualitatively analyse it in Figure \ref{fig: human_eval_qual}: AREL's story for the object-based example contains more monotonous sentences (`\textit{This is a picture of a city}) and  repetition between consecutive sentences. 

On contrary, our model can generate a more visually grounded and coherent story by utilising the storyline. While this example shows several useful concepts in the storyline that are not used in the generated story (`\textit{nativity scene}', `\textit{roman structure}'), concepts such as `\textit{tall}', `\textit{take picture}', and `\textit{tourists}' (highlighted in blue) did help in producing phrases related to these concepts, resulting in a story containing more interesting, diverse and relevant words. Furthermore, while there are error cases where the storyline contains irrelevant information such as the red words in the event-based example, this information was not included in the generated output. This is perhaps due to the advantages of the encoder-decoder cross-attentional mechanism of BART which allows the model to learn to select the more useful parts of the storyline.

Examining ReCo-RL, only the grounding and non-redundancy aspect received lower votes for object versus event-based instances. Compared to AREL, its better performance may be due to its framework incorporating RL rewards to directly align the outputs more to a human story in terms of the 3 criteria. PR-VIST however which first builds a storyline like ours, outperforms AREL and ReCo-RL and further, even yields slightly more votes for object-based stories compared to its proportion of votes received for event-based stories, thus highlighting the effectiveness of storyline and content planning. Despite PR-VIST's improvements, our approach and storyline construction method is evidently superior and substantially outperforms PR-VIST in all aspects across the 2 categories.

\section{Conclusion}
In this paper, we presented SCO-VIST, a multi-stage novel framework for visual storytelling that utilises social-interaction knowledge for enhancing commonsense reasoning in stories. We design a heterogeneous story graph with causal ordering that connects captions and commonsense extracted from external sources and employ shortest path algorithms to find the optimal storyline for story generation. Extensive experiments on the benchmark dataset, analysis of automatic metrics and human evaluations demonstrate that SCO-VIST outperforms existing baselines and is capable of generating diverse stories that are highly coherent with strong visual grounding.

\section*{Limitations}
\textbf{Benchmark Scope and Annotation} Due to the lack of a high-quality visual storytelling dataset, most recent studies on visual story generation use only one publicly available dataset, VIST. The dataset size is large enough but the dataset used in most visual storytelling research publications, including this study, was limited in scope. The VIST consists of images from Flickr, which is an image/video-based social media platform and includes mostly personal images that captures people's daily lives or events.
In addition, each Flickr album has 5 human written stories where each story is usually comprised of one sentence per image. Those human annotators are not the Flickr album owner and hence the gold standard annotations by annotators may not be perfectly matched with the intention of the original Flickr album. 
Future work should investigate how to mitigate this issue by establishing a new visual storytelling dataset via adopting the image album descriptions from the original authors, and providing better instructions for human annotators that map generated stories to objects/relations of images.

\textbf{Adaptability to Low-Resource Languages} Moreover, our model pipeline requires a pre-trained image captioning model in the first stage, which may not be available for low-resource languages that have relatively less data available for training natural language processing systems. The metrics used for evaluation are also only capable of judging English-written language. Nevertheless, our pipeline can be reproduced and future study should consider re-running experiments on other languages once models and data become available.

% lots of data processing
% only works for 5 images? 
% relies on pretrained models 
% no 2022 models to compare? 

% \section*{Ethics Statement}
% Scientific work published at EMNLP 2023 must comply with the \href{https://www.aclweb.org/portal/content/acl-code-ethics}{ACL Ethics Policy}. We encourage all authors to include an explicit ethics statement on the broader impact of the work, or other ethical considerations after the conclusion but before the references. The ethics statement will not count toward the page limit (8 pages for long, 4 pages for short papers).

\bibliography{anthology,custom}
\bibliographystyle{acl_natbib}
\clearpage

\appendix
\newpage
% \section{Qualitative Analysis}
% Figure \ref{fig: qual_analysis} presents two example stories for our SRL-pmi model versus 6 visual storytelling baseline models: AREL, GLACNet, KG-Story, ReCo-RL, PR-VIST and TAPM. 
% \begin{figure}[H]
%   \centering
%   \includegraphics[width=1\linewidth]{images/qual_analysis_5_baselines.drawio.png}
%   \caption{Generated stories for our SRL-pmi model versus the 6 baselines models. Blue/red words represent concepts relevant/irrelevant to the image sequence.}
%   \label{fig: qual_analysis}
% \end{figure}

\begin{table*}[t!]
  \centering
  \resizebox{1\linewidth}{!}{
  \begin{tabular}{@{}l||ccc|c|cc|c}
    \toprule
     \textbf{Model} & \textbf{RoViST-VG} & \textbf{RoViST-C} & \textbf{RoViST-NR} & \textbf{RoViST} & \textbf{UNION} & \textbf{SPICE} & \textbf{Story Len.} \\ \midrule
        CLIP-SRL-pmi & 69.6 & \underline{71.1} & \textbf{91.1} & \underline{77.3} & 68.5 & 10.8 & 49.5 \\
    BLIP-SRL-pmi & \underline{70.8} & 69.8 & 90.5 & 77.0 & \underline{72.1} & \underline{11.3} & \underline{51.2} \\
    VIST-SRL-pmi & \textbf{72.0} & \textbf{74.1} & \underline{90.6} & \textbf{78.9} & \textbf{82.5} & \textbf{12.5} & \textbf{57.6} \\
    \bottomrule
  \end{tabular}}
  \caption{RoViST, UNION, and SPICE scores recorded when using different captioning models. All models are implemented using the SRL-pmi SCO-VIST variant.} \label{caption_ablation_metrics} 
\end{table*}

\section{Caption Ablation Study}

\begin{figure}[H]
  \centering
  \includegraphics[width=1\linewidth]{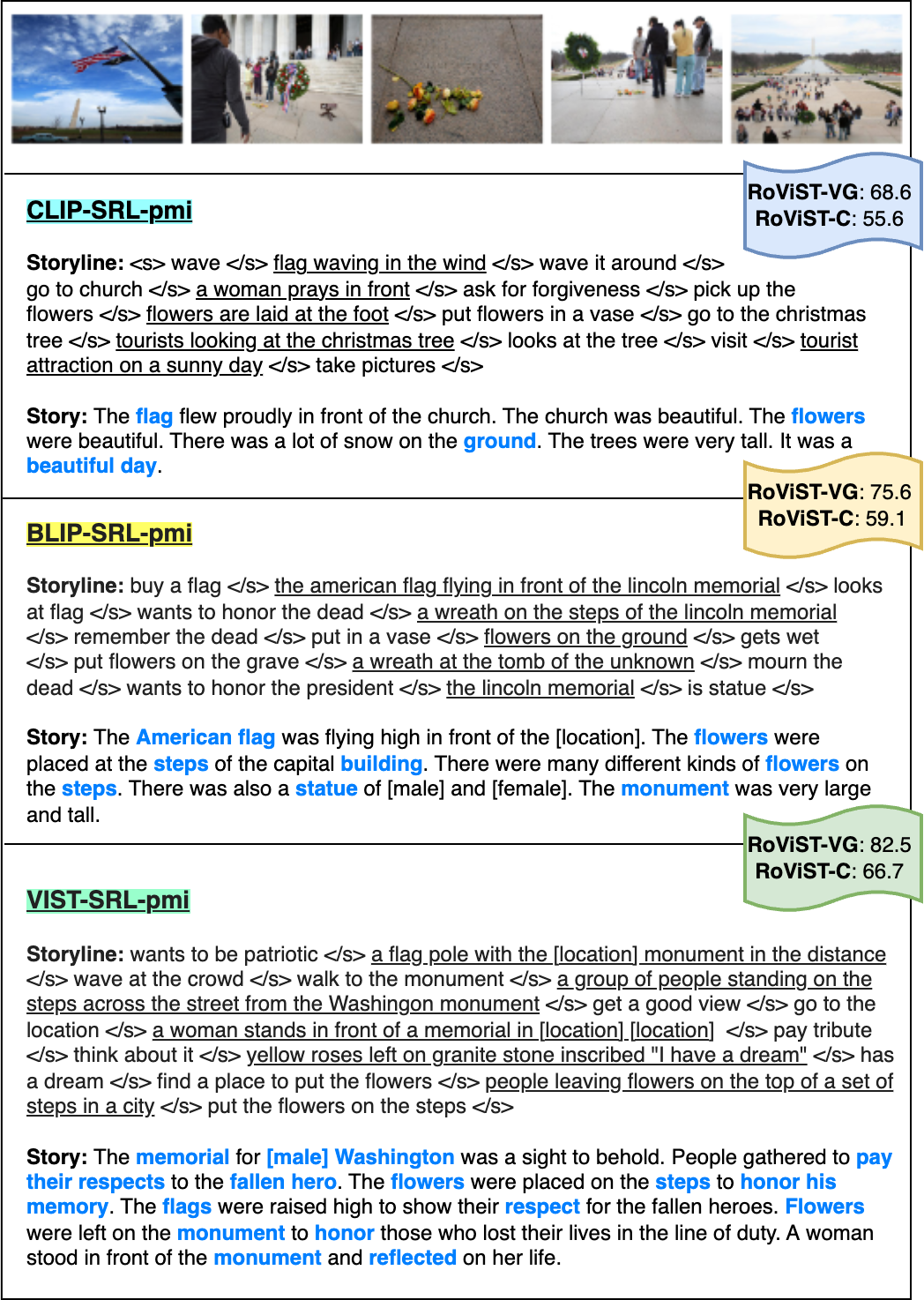}
  \caption{An example of a storyline and matching story generated using the SRL-pmi approach with different pre-trained image captioning models. Underlined words in the storyline are the image captions and blue words are visually relevant concepts to the image sequence.}
  \label{fig: caption_ablation}
\end{figure}

We conduct a preliminary ablation study to examine the performance of the stories when using different captioning models. For the experiments in the main paper, we utilised ClipCap \cite{mokady2021clipcap} to generate the image captions. For this experiment, we additionally consider the BLIP captioning model \cite{li2022blip} which outperforms ClipCap on COCO captions \cite{chen2015microsoft}. We also consider using the human-written captions which are provided as part of the VIST dataset. Note that for this experiment, we implement the SRL-pmi SCO-VIST variant for all models. Moreover, all models were trained on a substantially smaller dataset size (26939 instances for training, 3354 for validation and 3385 for testing) compared to the dataset used to retrieve the results in the main paper as ground-truth descriptions from VIST were only available for approximately half of the data. The CLIPCap and BLIP captions achieve a BLEU-1 score of 13.7 and 17.5 respectively when evaluated against the ground-truth VIST captions. 

The RoViST, UNION and SPICE scores using each captioning method is displayed in Table \ref{caption_ablation_metrics}. Firstly, it is evident that using human-written captions in the story graph creation process results in a higher RoViST-VG, RoViST-C and RoViST score overall as observed by VIST-SRL-pmi. UNION and SPICE were also considerably higher, suggesting better captions lead to better stories and SCO-VIST's outputs can be perhaps further improved with a stronger pre-trained captioning model. However for this study, we did find that using the BLIP captions produces a similar overall RoViST score. Neverthless, BLIP-SRL-pmi did yield greater RoViST-VG, UNION and SPICE compared to CLIP-SRL-pmi. The higher RoViST-VG score could imply that the caption quality influences the visual grounding aspect the most. This is reasonable as an incorrect caption could cause irrelevant concepts to be generated in the storyline, which can directly negatively impact the visual grounding score (RoViST-VG). % However, a story can still be coherent-sounding with little repetition even though it is not visually grounding. 

% This is reinforced by VIST-SRL-pmi's RoViST-VG score which achieves 72.0, higher than any score in Table \ref{automatic_metrics}, even though the model has been trained on significantly less data.

To highlight a specific example, we further conduct a qualitative analysis in Figure \ref{fig: caption_ablation} to assess how the caption quality can affect the generated storylines and stories. Taking CLIP-SRL-pmi for instance, the incorrect captions  \textit{`tourists looking at the christmas tree'} and \textit{`a woman prays in front'} results in irrelevant concepts mentioned in the story such as \textit{`church'} and \textit{`snow'}. Conversely, using more detailed and accurate captions as depicted in BLIP-SRL-pmi and VIST-SRL-pmi clearly results in better storylines which in turn, translates to more visually grounding and detailed stories. 
% There was no improvement in the coherence (RoViST-C) and the non-redundancy aspect (RoViST-NR), thus implying that caption quality influences the visual grounding aspect the most. This is reasonable as an incorrect caption could cause irrelevant concepts to be generated in the storyline, negatively impacting the visual grounding score (RoViST-VG).

% can be improved using a better captioning model but is beyond the scope of our work 

\section{Implementation Details} \label{implementation_details}
To generate the image captions for Stage 1, we use a pre-trained image captioning model called ClipCap \cite{mokady2021clipcap}. For commonsense generation, we use the `comet\_atomic2020\_bart' implementation of Comet-ATOMIC2020 \cite{hwang2021comet}. Sentence embeddings of the nodes are then obtained with a Sentence Transformer using the `all-mpnet-base-v2' model \cite{reimers-2019-sentence-bert} which outputs embeddings of size 768. Since some generated commonsense were found to be duplicated or similar, these similar or identical commonsense were filtered out based on if the sentence embedding cosine similarity score between the two phrases exceeded a threshold of 0.50 for each of the \texttt{BEFORE} and \texttt{AFTER} events produced by each caption.  

In Stage 2, the temporal GCN used to learn the node embeddings consisted of 1 layer and the chosen output dimension of the embeddings was 768. Furthermore, the Transformer model used to take in the 5 caption nodes to decode the story utilised the `bart-base' configuration of the BART Transformer model \cite{lewis2020bart}. This model was trained with a learning rate of 0.00001. 

In Stage 3, the story decoder using the storyline as input employed the `bart-large' configuration and was trained with a learning rate of 0.00002. For all BART models, we initialise with the pretrained weights and finetune them on our VST task. All experiments also used a batch size of 8, weight decay of 0.00001, learning rate decay of 0.95 scheduled to decrease after every epoch and the Adam optimizer \cite{kingma2015adam}. Early stopping was further employed to stop training after 3 consecutive epochs of no improvement on the validation set. At inference, we decode the story with nucleus sampling using the recommended values of $p$ = 0.9 and temperature = 0.9 \cite{holtzman2019curious}. All training of models was conducted using a Nvidia Tesla v100 16GB GPU which took approximately 15 hours to train. 
\\
\\
\textbf{Note that this is not the end of the Appendix section. The following page includes Appendix C, D, and E.}

\onecolumn

\section{Story Graph}\label{Story Graph}
\begin{figure}[t]
  \centering
  \includegraphics[width=1\linewidth]{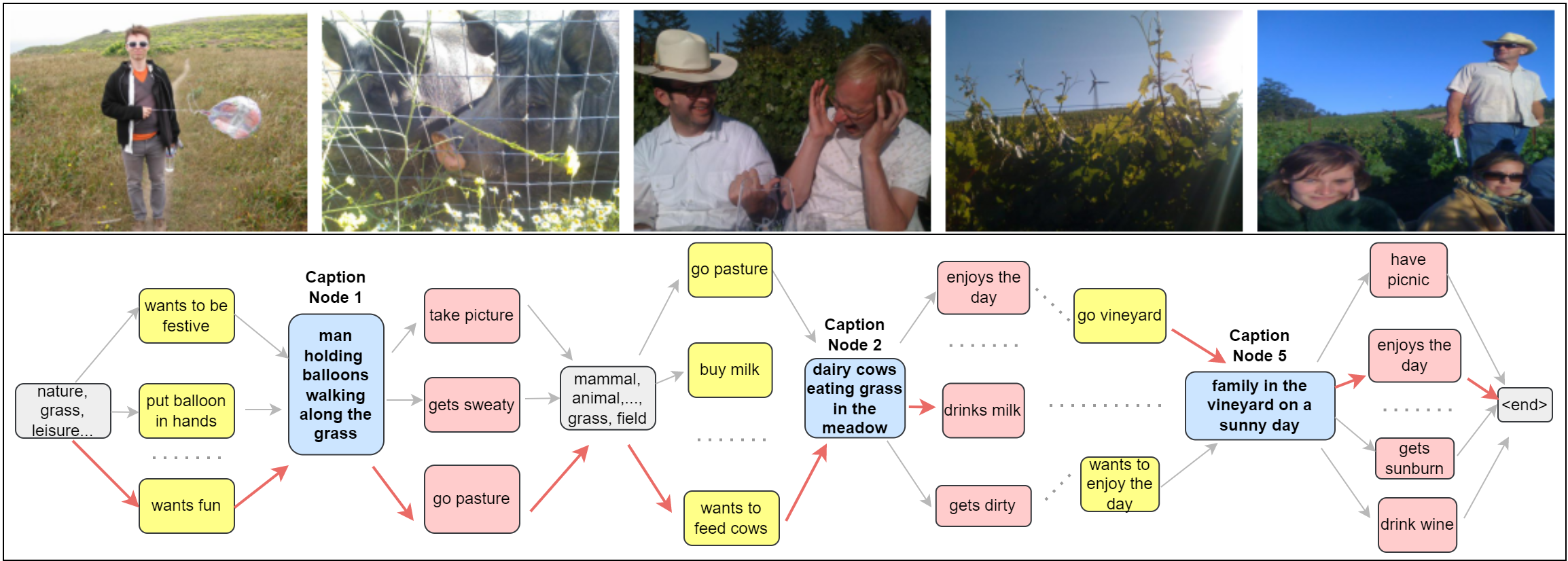}
  \caption{Final story graph generated from Stage 3 with red arrows indicating the optimal extracted storyline.}
  \label{fig: storygraph}
\end{figure}

\noindent Figure \ref{fig: storygraph} shows the final directed story graph generated from Stage 2 with the additional dummy end node added in Stage 3. Grey and blue nodes are theme and caption nodes respectively. Yellow nodes are commonsense nodes from the \texttt{BEFORE} events group generated by the \texttt{xNeed} and \texttt{xIntent} relation while red nodes are the \texttt{AFTER} events commonsense nodes from the \texttt{xWant} and \texttt{xEffect} relation. Due to limited space, only the nodes corresponding to image 1, 2 and 5 are visualised and dotted lines are used to indicate nodes in the graph that are not displayed. The red highlighted arrows show the shortest path found by Floyd Warshall's algorithm where the caption nodes and commonsense nodes are taken in order to use as the storyline. For simplicity, edge weights are also not shown.

\section{Human Evaluation Survey} \label{Appendix Survey}
\begin{figure}[H]
  \centering
  \includegraphics[width=1\linewidth]{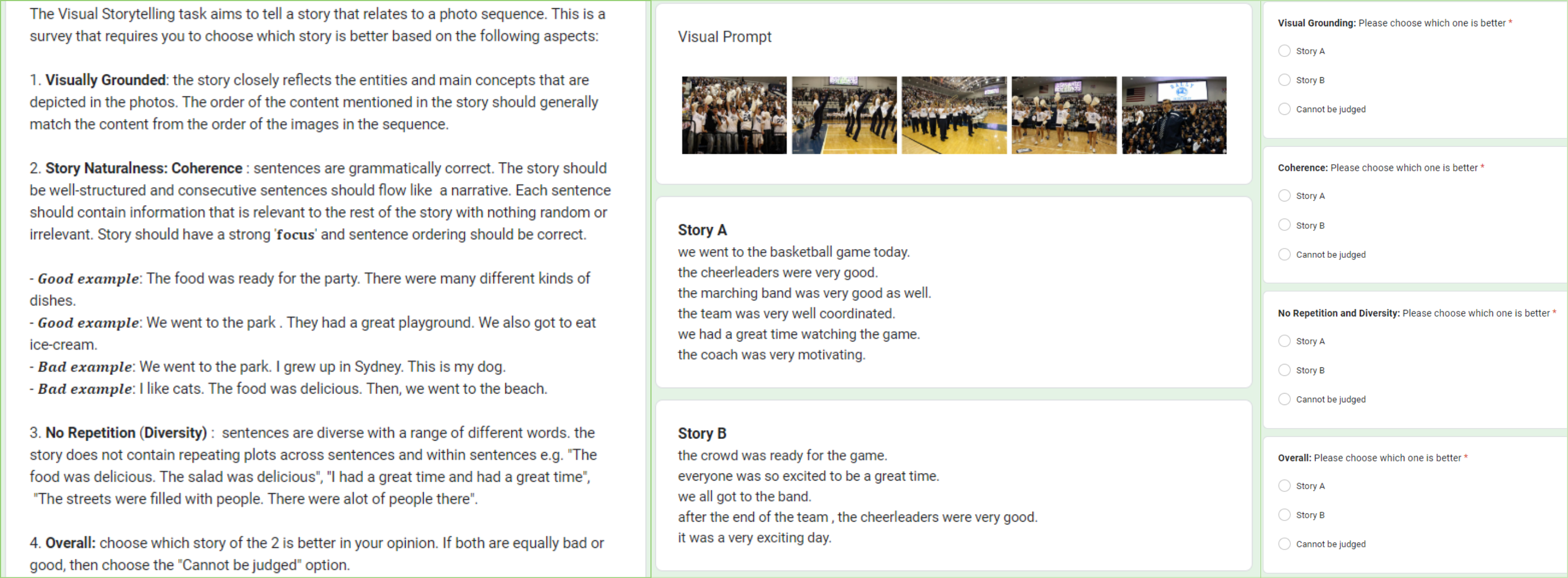}
  \caption{Survey instructions and form format for the human evaluation study.}
  \label{fig: survey}
\end{figure}

\noindent Figure \ref{fig: survey} shows the survey instructions used in the human evaluation study and the format of the survey questions. The 15 participants recruited were volunteers from a variety of age groups (20-60 years old), occupation and gender (8 female, 7 male). All participants were proficient in English with at least a university education level. Note that we modified and used similar instructions from the study proposed in \citet{wang-etal-2022-rovist}. It is also emphasised that annotators do not know which model generated which story as for each example, we randomly swap the order of the baseline story and SCO-VIST's story to be presented as Story A and Story B. 

\section{Event-based versus Object-based Stories}\label{event-object-appendix} 
\begin{figure}[H]
  \centering
  \includegraphics[width=1\linewidth]{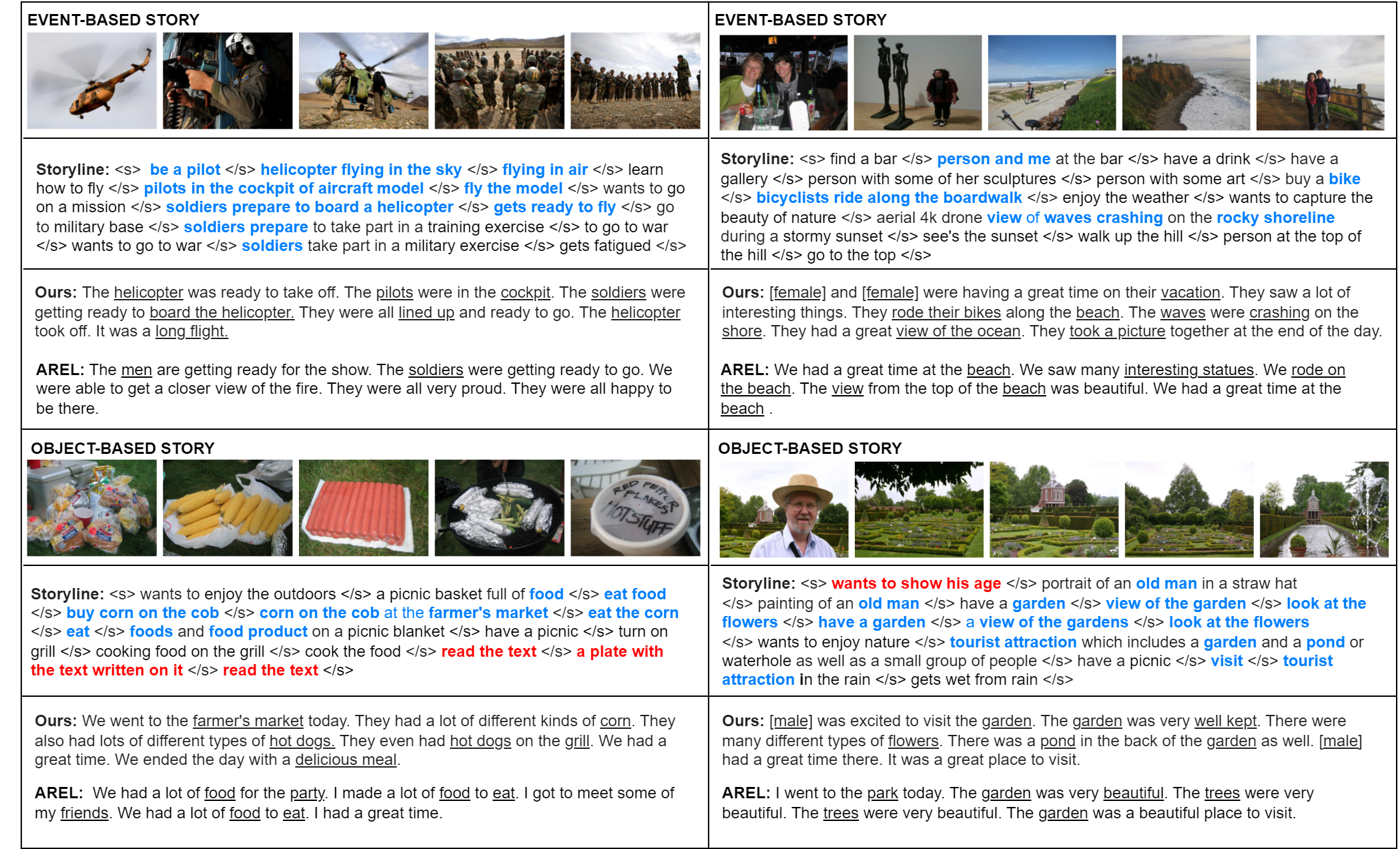}
  \caption{AREL versus our SRL-pmi model for event-based and object-based stories.}
  \label{fig: human_eval_qual_appendix}
\end{figure}

\noindent Figure \ref{fig: human_eval_qual_appendix} contains examples of more generated outputs from our SRL-pmi model versus AREL for event-based and object-based stories as described in Section 5.6 of the paper. Here, blue words in the storyline indicate concepts implicitly or explicitly used in the generated story while red words represent irrelevant or not useful concepts in the storyline. The underlined words in the generated story represent concepts relevant to the image stream.

\end{document}